\documentclass[12pt,a4paper,]{article}

\usepackage[utf8]{inputenc}
\UseRawInputEncoding
\usepackage[english]{babel}
\usepackage{graphicx}
\usepackage{amsmath}
\usepackage{amssymb}
\usepackage{booktabs}
\usepackage{caption}
\usepackage{subcaption}
\usepackage[margin=1in]{geometry}
\usepackage{hyperref}
\usepackage{float}
\usepackage{multirow}
\usepackage{array}
\usepackage{textcomp} 
\usepackage[numbers]{natbib}
\usepackage{xcolor}

\graphicspath{{figures/}}

\title{From Papers to Property Tables: A Priority-Based LLM Workflow for Materials Data Extraction}
\author{Koushik Rameshbabu$^{1*}$, Jing Luo$^2$, Ali Shargh$^2$, Khalid A. El-Awady$^2$, \\
Jaafar A. El-Awady$^{2\ddagger}$\\
$^1$Department of Applied Mathematics and Statistics, Johns Hopkins University\\
$^2$Department of Mechanical Engineering, Johns Hopkins University\\
\texttt{$^*$kramesh5@jh.edu, $^\ddagger$jelawady@gmail.com}}
\date{\today}

\begin{document}

\maketitle

\begin{abstract}
Scientific data are widely dispersed across research articles, often reported inconsistently across text, tables, and figures, making manual data extraction and aggregation slow and error-prone. We present a prompt-driven, hierarchical workflow that uses a large language model (LLM) to automatically extract and reconstruct structured, shot-level shock-physics experimental records by integrating information distributed across text, tables, figures, and physics-based derivations from full-text published research articles, using alloy spall strength as a representative case study. The pipeline targeted 37 experimentally relevant fields per shot and applied a three-level priority strategy: (T1) direct extraction from text/tables, (T2) physics-based derivation using verified governing relations, and (T3) digitization from figures when necessary. Extracted values were normalized to canonical units, tagged by priority for traceability, and validated with physics-based consistency and plausibility checks. Evaluated on a benchmark of 30 published research articles comprised of 11,967 evaluated data points, the workflow achieved a high overall accuracy, with priority-wise accuracies of 94.93\% (T1), 92.04\% (T2), and 83.49\% (T3), and an overall weighted accuracy of 94.69\%. Cross-model testing further indicated strong agreement for text/table and equation-derived fields, with lower agreement for figure-based extraction. Implementation through an API interface demonstrates the scalability of the approach, achieving consistent extraction performance and, in a subset of test cases, matching or exceeding chat-based accuracy.
This workflow demonstrates a practical approach for converting unstructured technical literature into traceable, analysis-ready datasets without task-specific fine-tuning, enabling scalable database construction in materials science.
\end{abstract}

\section{Introduction}

Recent advances in machine learning and artificial intelligence have underscored the critical importance of both data quality and quantity in many domains of science and physics, thus necessitating the development of comprehensive training databases. In the field of materials science and engineering, there are a limited number of standard databases \cite{belsky_new_nodate, allen_iucr_nodate, noauthor_oqmd_nodate}; however, much data on materials remains fragmented, which presents challenges for sharing, comparison, and integration between various research groups \cite{seshadri_perspective_2016,Kalidindi2016Role, Ciardi2024Materials, Bauer2024Roadmap}. This need has become even more acute as data-centric materials science and AI-driven discovery increasingly depend on accessible, structured, and provenance-preserving datasets assembled from published literature \cite{seshadri_perspective_2016,Bauer2024Roadmap}. This fragmentation not only impedes collaborative research and the advancement of knowledge, but also increases the risk of redundancy and errors, as researchers may inadvertently duplicate efforts or misinterpret data due to inconsistent reporting standards. To effectively address these challenges, it is essential to develop databases from existing literature that are specifically customized to meet the requirements of materials science.
 
The construction of databases in materials science presents significant challenges that require not only a deep understanding of materials science knowledge but also the capability to integrate diverse data sources and experimental results. The heterogeneity of data types, including both experimental measurements and computational simulations, coupled with decades of inconsistent terminology, symbols, and units across subfields, publication eras, and geographic origins of research, complicates the systematic extraction of data. Furthermore, data may be embedded in text, tables, or figures within a document, and the information needed for a single record may be distributed across all three. 

Prior materials literature-mining approaches instead are based on rule-based natural language processing (NLP), manual curation, domain-specific transformer models, and, more recently, prompt-based large language models (LLM) extraction. Nevertheless, rule-based pipelines and hand-built grammars can be brittle and labor-intensive to adapt to new domains or reporting styles \cite{han_successful_2015,si_research_2020}, manual curation can result in high-quality datasets, but do not scale to large volumes of documents \cite{chen_fatigue_2022,gorsse_dataset_2023}, and transformer-based workflows generally require labeled training data and task-specific fine-tuning for optimal performance \cite{choi_quantitative_2024,foppiano_mining_2024}. Materials-domain models such as MatSciBERT have improved named-entity recognition and relation classification \cite{Gupta2022}, while BERT-PSIE demonstrated rule-free text mining for automated compound-property database construction \cite{Gilligan2023}. However, these approaches are generally designed for specific extraction tasks (e.g., entity recognition or property relations), often operate primarily on text, and typically require task-specific training data or pipeline customization. As a result, such approaches may not be readily extensible to reconstruct complete experimental records when information is distributed across text, tables, and figures. These remaining challenges motivated the development of more flexible, prompt-driven workflows capable of integrating heterogeneous data sources and domain-specific reasoning within a unified framework.  



Recent advances in LLMs offer promising solutions to different challenges in materials research (Cf. \cite{buehler_mechgpt_2024,chandrasekhar_amgpt_2024,nanogpt_chandrasekhar_2025,choudhary_microscopygpt_2025,Choudhary2024}). With regards to data extraction from literature, ChatExtract showed that conventional LLMs, combined with prompt engineering and follow-up validation, can extract high-quality materials data from research articles \cite{polak_extracting_accurate_materials_2024}, while Foppiano et al. evaluated LLMs on named-entity and relation-extraction tasks in materials science \cite{foppiano_mining_2024}. Additional recent studies have used LLM-assisted workflows to build databases of organic and inorganic materials from the literature, indicating that prompt-based extraction can support scalable database construction with substantially less task-specific coding than many earlier approaches \cite{hu_ai_powered_workflow_2025,liu_expert_grounded_prompt_2025}. 

It should be noted that most prior LLM-based extraction approaches focused on text-only extraction, table-centric extraction, entity/relation extraction, or property-specific database construction rather than the reconstruction of complete experimental records. This gap is important because materials data are often distributed across text, tables and figures. Table-extraction work, such as DiSCoMaT, and multimodal benchmarks, such as MatViX, make clear that extracting structured information from visually rich, full-length materials articles remains challenging for current systems \cite{Gupta2023,Khalighinejad2025}.



These limitations are especially evident in shock-physics literature, where a single paper may report many shot-level experiments, with some variables explicitly tabulated, others embedded in narrative text, others recoverable only through governing physical relations, and still others available only in plots. Here, we demonstrate a structured, LLM prompt-driven workflow for automated extraction and reconstruction of shot-level shock-physics data from scientific research articles, using alloy spall strength as a representative case study. The contribution is therefore not the extraction of isolated entities or property-value pairs, but the reconstruction of complete experimental records from heterogeneous reporting formats.

Our approach is centered on a hierarchical three-level priority strategy in the prompt: it first extracts explicitly reported values from text and tables, then derives missing quantities using verified shock-physics relationships, and finally resorts to figure digitization only when necessary. Extracted values are then normalized to canonical units, tagged by extraction priority for provenance, and validated with physical-consistency and plausibility checks. This structure mirrors the decision logic of expert curators and enables traceable, physically consistent extraction without model fine-tuning. Instead of relying on ad hoc rules or post-processing scripts, the proposed workflow integrates domain knowledge, logical constraints, and verification steps into a multi-level prompting framework. This enables the LLM to consistently, traceably, and physically extract experimental data from complex technical papers.

The paper is organized as follows. Section~\ref{method} outlines the methodology, detailing the hierarchical extraction priorities and the prompt-guided workflow. Section~\ref{results} applies this approach to papers on high strain loading of alloys, serving as a representative case study, and presents quantitative performance metrics and validation results. Section~\ref{discussion} explores the broader implications and limitations, while Section~\ref{conclusion} provides concluding remarks.


\section{Methodology}\label{method}
\subsection{System Overview and Flow}
The data extraction approach was implemented as an end-to-end pipeline powered by a large language model (LLM). Figure 1 shows a flow chart of this pipeline, demonstrating how documents are processed to generate a structured dataset \cite{knime_llm_toxicology_2025}. The pipeline consists of an LLM (Gemini 3 Pro unless otherwise stated) that reads the document content (including all text, tables, and figures) and generates a predefined table of user-defined parameters. 

After each document is ingested and parsed by the Vision language model (VLM), the workflow converts the content into model-readable inputs (e.g., extracted text, tables, and figure panels) and extracts user-defined parameters. In the current study, we define 37 parameters commonly considered highly relevant for characterizing the spall strength in metals deformed at high strain rates. These 37 parameters are listed in Table~\ref{tab:fields_37_sno}. Each target field is obtained from the most reliable available source according to the following extraction track priority: (i) extraction from text and/or tables (Tier 1), (ii) derivation using prescribed equations when sufficient inputs are available (Tier 2), and (iii) extraction from figures (Tier 3). It should be noted that Tier 2 is prioritized over Tier 3 because equation-based derivations use governing relationships and given inputs, whereas figure-based extraction can introduce digitization uncertainty (e.g., due to axis resolution, marker ambiguity, and scaling), especially from older documents. The outputs are then consolidated into a single standardized table, with each entry tagged with its extraction priority (e.g., Tier 1/Tier 2/Tier 3) to preserve traceability and enable for downstream audit and validation \cite{li_dual_llm_adversarial_2025}.

\begin{figure}[H]
    \centering
    \includegraphics[width=0.88\textwidth]{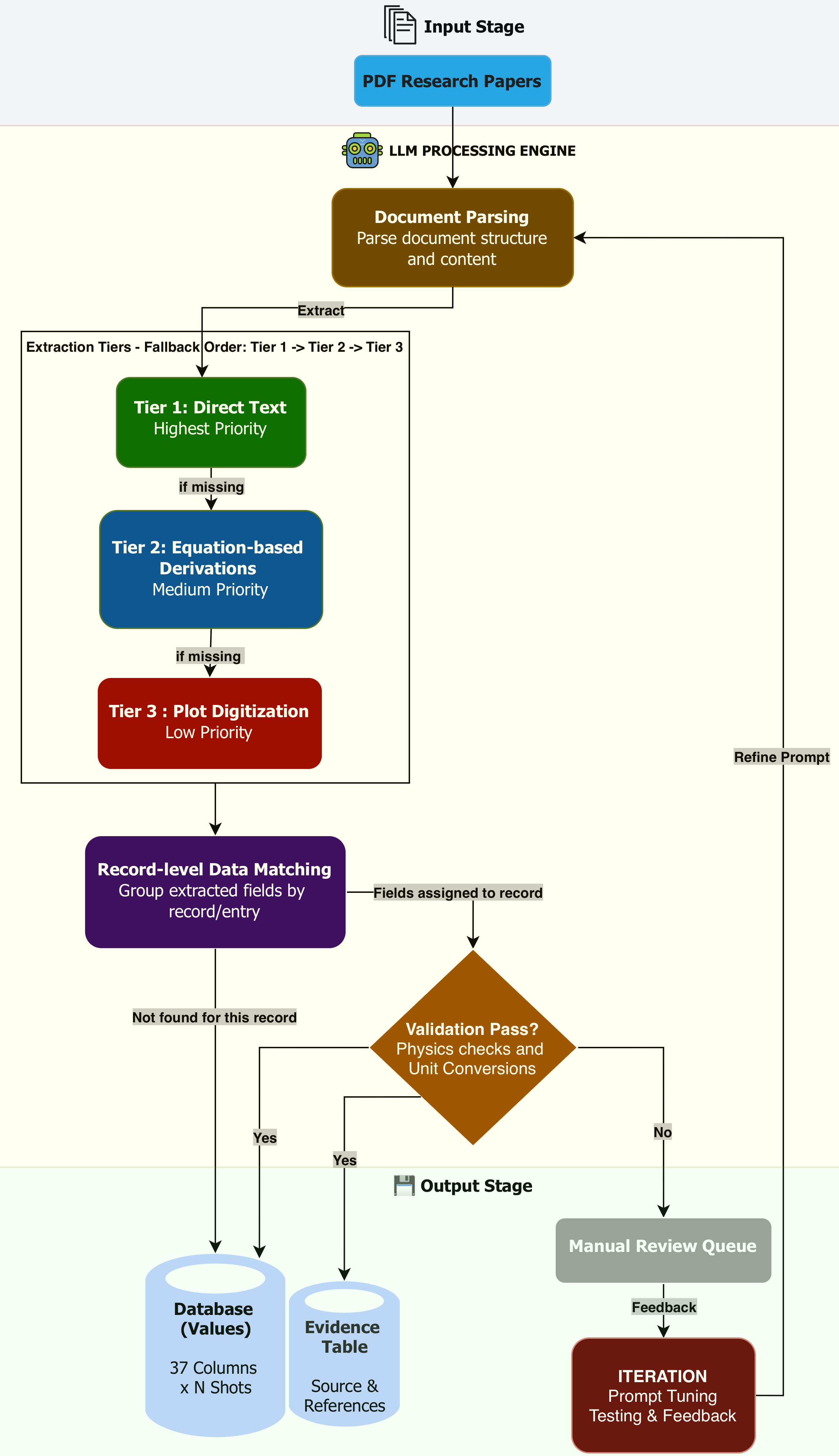}
    \caption{Prompt-development and refinement workflow for LLM-based data extraction from documents.}
    \label{fig:pipeline}
\end{figure}

As shown in Figure~\ref{fig:pipeline}, we also incorporate a post-consolidation validation step. After values are added to the extracted dataset, the pipeline (i) normalizes all units using a consistent convention defined by the user, (ii) performs basic physical plausibility and internal-consistency checks, and (iii) flags anomalies for follow-up (e.g., missing mandatory fields, unit mismatches, or physically implausible combinations of parameters). Entries that pass these checks are written to the final structured dataset, while flagged cases are routed to a manual-review queue to preserve data fidelity. This verification stage serves as a quality-control gate that helps ensure that the exported database is both analysis-ready and comparable across documents \cite{hu_ai_powered_workflow_2025}.

\begin{table}[H]
\centering
\small
\caption{List of the 37 extracted fields defining the spall-strength dataset schema.}
\label{tab:fields_37_sno}
\renewcommand{\arraystretch}{1.25}
\setlength{\tabcolsep}{8pt}
\begin{tabular}{p{0.68\textwidth} p{0.18\textwidth}}
\toprule
\textbf{Extracted Field} & \textbf{Symbol} \\
\midrule
Metal Symbol & -- \\
Sample ID & -- \\
Synthesis Method & -- \\
Treatment & -- \\
Initial Temperature (K) & $T_0$ \\
Quasi-static Yield Stress (MPa) & $\sigma_y$ \\
Free Surface Velocity at Hugoniot Elastic Limit (HEL) (m/s) & $u_{\mathrm{HEL}}$ \\
Shear Stress at HEL (GPa) & $\tau_{\mathrm{HEL}}$ \\
Hardness & -- \\
Bulk Modulus (GPa) & $B$ \\
Shear Modulus (GPa) & $G$ \\
Young's Modulus (GPa) & $E$ \\
Poisson's Ratio & $\nu$ \\
Melting Point (K) & $T_m$ \\
Sample Thickness (mm) & -- \\
Sample Diameter (mm) & -- \\
Grain Size (\textmu m) & -- \\
Initial Density (g/cm$^{3}$) & $\rho_0$ \\
Longitudinal Sound Speed (m/s) & $c_L$ \\
Shear Sound Speed (m/s) & $c_s$ \\
Bulk Sound Speed (m/s) & $c_b$ \\
Flyer Material Name & -- \\
Flyer Material Code & -- \\
Flyer Thickness (mm) & -- \\
Flyer Diameter (mm) & -- \\
Impact Velocity (m/s) & $v_i$ \\
Longitudinal Stress at HEL (GPa) & $\sigma_{\mathrm{HEL}}$ \\
Peak Stress (GPa) & $\sigma_{\mathrm{peak}}$ \\
Strain Rate (s$^{-1}$) & $\dot{\varepsilon}$ \\
Pulse Duration (\textmu s) & -- \\
Experiment Type & -- \\
Gas Gun Diameter (mm) & -- \\
Spall Strength (GPa) & $\sigma_{\mathrm{sp}}$ \\
Spall Pullback Velocity (m/s) & $\Delta u_{\mathrm{pb}}$ \\
Reference Title & -- \\
DOI & -- \\
Verification & -- \\
\bottomrule
\end{tabular}
\end{table}

To guide the LLM in the extraction of data, we imposed domain-specific criteria for unit conversion, physical consistency, and output format \cite{polak_extracting_accurate_materials_2024}. As summarized in Table~\ref{tab:prompt_rules}, the prompt was divided into five distinct sections that included: (i) interpretation and normalizations (rules A--B), which resolves symbols to physical definitions and preserves uncertainty notation; (ii) extraction strategy and priority (rules C--E), which enforces the priority hierarchy in extracting data from text, tables, and and figures when necessary; (iii) field-specific equation-based extraction rules (rules F--G), which apply unit conversions and defines governing relations; (iv) output specifications (rule H); and (v) provenance and execution checks (rules I--J).
This structured prompt design provides the LLM with comprehensive and sequential instructions that constrain it to adhere to domain-specific criteria for unit conversion, extraction priority, physical consistency, and output format, thereby enabling more accurate data extraction. The complete information for each section is provided in Supplementary Materials Section S1.2\cite{supplement2026}.

\begin{table}[H]
  \centering
  \small
  \caption{Operational rules for guiding the data extraction process, grouped by class.}
  \label{tab:prompt_rules}
  \renewcommand{\arraystretch}{1.15}
  \begin{tabular}{>{\raggedright\arraybackslash}p{0.20\textwidth}
                  >{\raggedright\arraybackslash}p{0.26\textwidth}
                  >{\raggedright\arraybackslash}p{0.48\textwidth}}
    \toprule
    \textbf{Class} & \textbf{Rule Name} & \textbf{Description} \\
    \midrule
    \multirow{2}{=}{Interpretation \& Normalization}
      & A. Symbol Disambiguation
      & Resolves ambiguous symbols to physical quantities \\
      & B. Uncertainty Extraction
      & Preserves original error formats (e.g., $\pm$, parenthesis) \\
    \midrule
    \multirow{3}{=}{Extraction Strategy \& Priority}
      & C. Data Priority
      & Define prioritization of extraction sources: Direct $\rightarrow$ Calculated $\rightarrow$ Figure \\
      & D. Record Identification and Grouping
      & Identify and extract parameters from text or tables \\
      & E. Figure Extraction
      & Apply centroid logic to digitize data points from figures \\
    \midrule
    \multirow{2}{=}{Physics Constraints \& Derivations}
      & F. Unit Conversion
      & Enforce canonical units across all output fields \\
      & G. Physics-Based Derivations
      & Define the physics-based formulas (e.g., stress equations for $\sigma_{\text{HEL}}$, $\sigma_H$, and $\tau_{\text{HEL}}$, as well as elastic moduli derivations from wave speed equations) \\
    \midrule
    Output
      & H. Output Specification
      & Define the specifications for the dataset \\
    \midrule
    \multirow{2}{=}{Provenance \& Execution Checks}
      & I. Evidence Log Specification
      & Define the structure of the supporting evidence for validation \\
      & J. Execution Checklist
      & Validate final consistency across all entries \\
    \bottomrule
  \end{tabular}
\end{table}

A central component of the prompt is a three-tier priority hierarchy that instructs the LLM to select the most reliable available source for each field and follow a consistent decision path across different runs. This priority-based prompt reduces output variability by preventing the LLM from switching extraction strategies between executions. In this strategy, Tier 1 targets direct extraction from text or tables; Tier 2 derives missing quantities using governing relations when sufficient inputs are already available; and Tier 3 performs figure-based extraction only if Tiers 1 and 2 do not identify a field value. Tier 3 has the lowest priority to avoid uncertainties associated with plot digitization. Additionally, if a value is extracted from Tier 1, it is retained as authoritative; values from Tier 2 are used only when Tier 1 does not produce a value or to validate the extraction from Tier 1. If neither Tier 1 nor Tier 2 result in a value, the workflow proceeds to Tier 3 when applicable. Finally, if both Tiers 1 and Tier 2 extract two values that differ by more than a user-predefined tolerance, the entry is flagged for manual review, and both sources are recorded in the evidence log.


Tier 1 extraction is the highest-priority level because it is based on explicitly reported values, which minimizes interpretation and compounding errors. The model extracts values directly from the text or tables of a paper, including preserving any associated uncertainty (e.g., $3730\pm 20\,\mathrm{m/s}$). 
The prompt also includes explicit instructions to prevent the LLM from hallucinating data and to ensure that if a Tier 1 value is missing, it is left blank (``--'') in the dataset. The model is also instructed to preserve the reported uncertainty notation (e.g., $\pm$ values or parenthetical notation) rather than re-interpreting it.


If a required value cannot be found via the Tier 1 extraction, the LLM proceeds to Tier 2, where missing fields are derived from other reported quantities using established shock-physics relations provided in the prompt. This level is ranked below Tier 1 because it relies on computations beyond direct reporting, but above Tier 3 because the derivations are anchored in explicit inputs and well-defined physical laws. The prompt includes a library of governing equations, which are summarized in Table~\ref{tab:equations}. 

These physics-based calculations are mandated when possible. For example, the prompt's ``mandatory backtracking'' rule requires that if certain inputs are present (e.g., $\sigma_{HEL}$, density $\rho_0$, and wave speeds), the model must derive the missing complementary values rather than leaving them blank. It is interesting to note that in some cases the LLM found additional relevant equations in the manuscripts that were not explicitly provided in the prompt and utilized them to compute the some of the missing data. For example, in the article by Hillel et al. (2022)~\cite{hillelShockWaveStudy2022}, the LLM was able to identify the strain-rate equation, extract the required parameters from the shot tables, and perform the necessary matching and calculations, although that particular formula was not defined in the prompt. 

All Tier 2-derived entries are marked as ``calculated'' in the model's internal notes for traceability. Finally, to maintain physical realism, the prompt also provided sanity-check ranges (e.g., expecting $\tau_{HEL}/\sigma_{HEL}\approx 0.2$--$0.35$ and $c_s/c_L\approx 0.5$--$0.6$ as typical values for metallic systems~\cite{thomasDynamicStrengthProperties2018}) to ensure the model flags implausible calculations. 

\begin{table}[H]
    \centering
    \small
    \caption{Physics-based relationships utilized for Tier 2 calculations.}
    \label{tab:equations}
    \renewcommand{\arraystretch}{2.0}
    \begin{tabular}{p{0.40\textwidth}p{0.50\textwidth}}
        \toprule
        \textbf{Parameter} & \textbf{Formula (in SI units)} \\
        \midrule
        Longitudinal stress at HEL & 
        $\displaystyle \sigma_{\text{HEL}} = 0.5\,\rho_{0}\, c_{L}\, u_{\text{HEL}}$ \\[0.5em]
        
        Shear stress at HEL & 
        $\displaystyle \tau_{\text{HEL}} = \left( \frac{c_{s}}{c_{L}} \right)^{2}\,\sigma_{\text{HEL}}$ \\[0.5em]
        
        Spall strength & 
        $\displaystyle \sigma_{\text{sp}} = 0.5\,\rho_{0}\, c_{b}\,\Delta u_{\text{pb}}$ \\[0.5em]
        
        Bulk sound speed & 
        $\displaystyle c_{b} = \sqrt{\, c_{L}^{2} - \frac{4}{3}c_{s}^{2}\,}$ \\[0.5em]
        
        Shear modulus & 
        $\displaystyle G = \rho_{0}\, c_{s}^{2}$ \\[0.5em]
        
        Bulk modulus & 
        $\displaystyle B = \rho_{0}\, c_{b}^{2}$ \\[0.5em]
        
        Longitudinal modulus & 
        $\displaystyle M = \rho_{0}\, c_{L}^{2}$ \\[0.5em]
        
        Young's modulus & 
        $\displaystyle E = \frac{9BG}{3B + G}$ \\[0.5em]
        
        Poisson's ratio & 
        $\displaystyle \nu = \frac{3B - 2G}{6B + 2G}$ \\[0.5em]
        
        Pullback velocity & 
        $\displaystyle \Delta u_{\text{pb}} = \frac{2\sigma_{\text{sp}}}{\rho_{0}c_{b}}$ \\[0.5em]
        
        Free surface at HEL & 
        $\displaystyle u_{\text{HEL}} = \frac{\sigma_{\text{HEL}}}{0.5\rho_{0}c_{L}}$ \\[0.5em]
        \bottomrule
    \end{tabular}
\end{table}

\begin{figure}[H]
    \centering
    \includegraphics[width=0.8\textwidth]{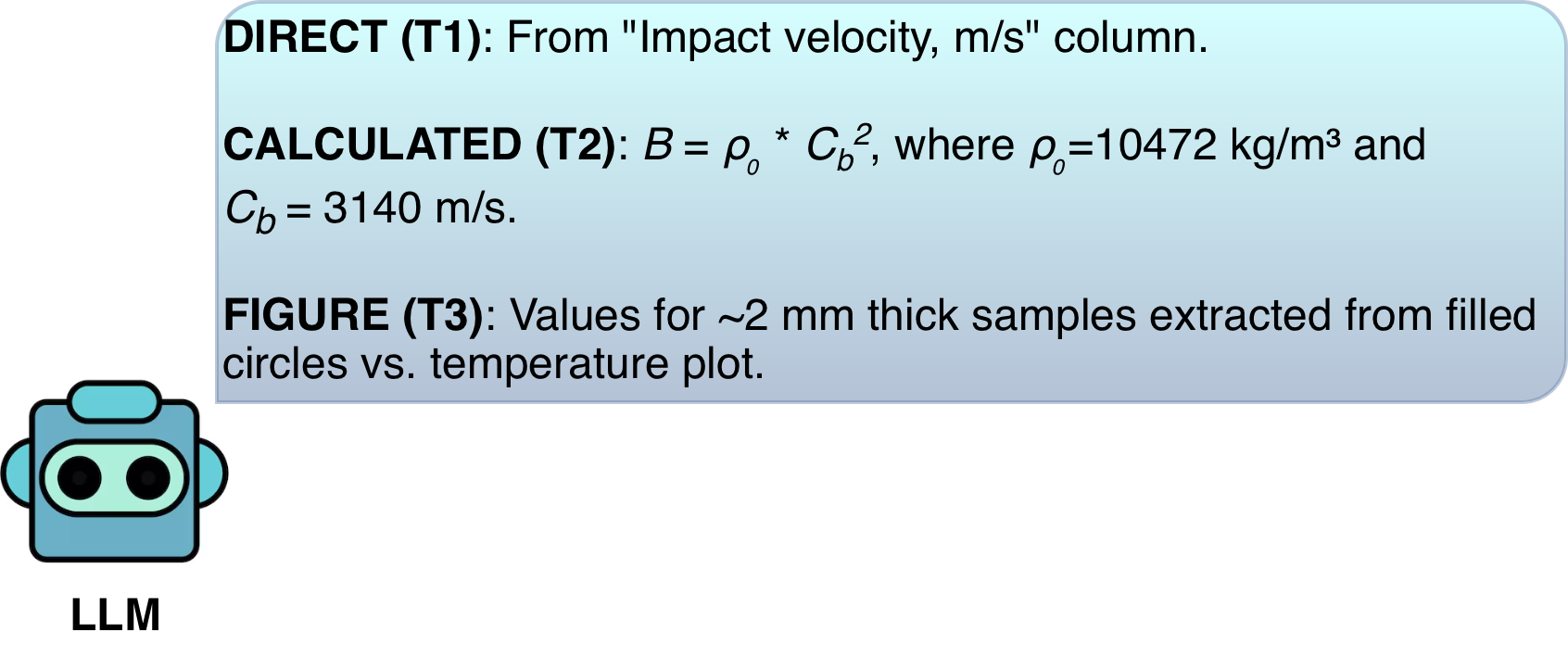}
    \caption{Example evidence log for Tier 1 (direct), Tier 2 (calculated), and Tier 3 (figure) extractions.}
    \label{fig:priority_examples}
\end{figure}

Figure~\ref{fig:priority_examples} shows representative evidence logs for all three extraction tiers and illustrates how the provenance of each extracted field is documented throughout the pipeline. In the Tier 2 example, the bulk modulus, $B$, is not explicitly reported in the article, so it is computed from other variables reported in the document using the appropriate physics-based laws provided in the prompt. The corresponding evidence log also reports the quantities and equations used in the derivation, ensuring the calculation remains fully traceable. The figure also includes a Tier 3 example of figure-based extraction, where spall-strength values for 2 mm thick samples are obtained from the filled-circle markers in the temperature-dependency plot in the paper. This example shows that the process involves more than simply reading values from a graph. It also requires matching the plot data-points to the correct experimental subset so that only the relevant shot entries are populated, while the remaining entries are left blank. Taken together, these examples show that the evidence-log framework supports transparency, preserves the origin of each extracted value, and enables downstream verification of whether a field was obtained directly from text, calculated from reported variables, or digitized from a graphical source.



\subsection{Evaluation Metrics and Accuracy Definitions}

Categorical fields, including material, condition, and experimental record ID, were scored as correct only if they exactly matched the ground truth after normalization of whitespace and capitalization. The ground truth was constructed from manually curated values extracted from the same document, cross-verified with existing databases where available, and complemented by manually digitized values for graph-derived quantities. For numerical fields, values were converted to canonical units and scored as correct if they matched the ground truth within a relative tolerance of less than $0.5\%$, according to $|x - x_{\mathrm{gt}}| / \max(|x_{\mathrm{gt}}|,\epsilon) < 0.005$, where $\epsilon = 10^{-12}$ was introduced to prevent division by values close to zero. The $0.5\%$ threshold was chosen as a practical compromise. It is small enough to detect meaningful extraction or derivation errors, but large enough to accommodate minor discrepancies arising from rounding, unit conversion, and manual graph digitization. Missing entries were counted as correct only when the corresponding value was genuinely absent from the source, as confirmed during manual ground-truth curation.

For a document indexed by $p$  and priority level $k \in \{\text{T1, T2, T3}\}$ (which corresponds to Tier 1, 2, and 3, respectively), the priority level accuracy is:
\begin{equation}
    \text{Acc}_{p,k} = \frac{N_{\text{correct},p,k}}{N_{\text{total},p,k}},
\end{equation}
where $N_{\text{correct},p,k}$ is the number of correctly extracted fields with priority $k$ for paper $p$, and $N_{\text{total},p,k}$ is the number of fields evaluated with priority $k$ for document $p$ \cite{shi_llm_mof_extraction_2025}.

For document $p$, the weighted accuracy across all priorities is defined as:
\begin{equation}
    \text{Acc}_{p, \text{weighted}} = \frac{\sum_{k \in \{\text{T1, T2, T3}\}}N_{\text{total}, p, k}\, \text{Acc}_{p, k}}{\sum_{k \in \{\text{T1, T2, T3}\}}N_{\text{total}, p, k}},
\end{equation}

Equivalently:
\begin{equation}
    \text{Acc}_{p, \text{weighted}} = \frac{\sum_{k \in \{\text{T1, T2, T3}\}}N_{\text{correct}, p, k}}{\sum_{k \in \{\text{T1, T2, t3}\}}N_{\text{total}, p, k}},
\end{equation}


The overall weighted accuracy across all papers is computed as follows:

For $P$ number of documents, the overall weighted accuracy is:
\begin{equation}
    \text{Acc}_{\text{overall}} = \frac{\sum_{p=1}^{P}{\sum_{k \in \{\text{T1, T2, T3}\}}N_{\text{correct}, p, k}}}{\sum_{p = 1}^{P}{\sum_{k \in \{\text{T1, T2, T3}\}}N_{\text{total}, p, k}}}
\end{equation}

Additionally, to quantify performance for each priority tier across all documents, we also computed the overall weighted accuracy for each priority:
\begin{equation}
    \text{Acc}_{\text{overall}, k} = \frac{\sum_{p = 1}^{P}N_{\text{correct}, p, k}}{\sum_{p = 1}^{P}N_{\text{total}, p, k}} \quad\quad k \in \{\text{T1, T2, T3}\}
\end{equation}

\section{Results}\label{results}

\subsection{Overall Extraction Performance}

Using the methodology described above, we first extracted structured data through the Gemini 3 Pro's chat interface from 30 articles published between 1996 and 2024~\cite{ameriSpallStrengthDependence2022, boddorffSpallFailureAdditively2022, chenSpallBehaviorAluminum2006, chengShockCompressionSpallation2022, cottonSpallStrengthNiobium2012, dandekarShockResponseHeavy, farbaniecSpallResponseFailure2017, farbaniecMicrostructuralEffectsSpall2016, fensinDynamicFailureTwophase2015, grayStructurePropertyConstitutive2017, hawkinsDynamicPropertiesFeCrMnNi2022, hillelShockWaveStudy2022, jiaoPhaseTransitionTwinning2023, kanelSpallFractureProperties1996, liSpallDamageMild2016, liSpallStrengthMild2019, luShockSpallationBehavior2023, maEffectsAlloyingElement2019, millettBehaviorNiNi60Co2008, neelShockSpallLowalloy2020, songPlasticDeformationBehavior2024, thomasDynamicStrengthProperties2018, whelchelSpallDynamicYield2014, williamsSpallResponse1100O2012, xieHydrogenInducedSlowdown2021, yangEffectsPhaseContent2021, zaretskyImpactResponseCobalt2010, zaretskyImpactResponsePrestrained2022, zaretskyPlasticFlowShockloaded2011, zhangShockCompressionSpallation2022}. The common focus of all these articles is on experimentally measuring the spall strength of different metallic alloys. Performance was evaluated by comparing the LLM-generated tables against ground-truth data consisting of manually curated values from the same articles, cross-verified with existing databases, and manually digitized graph values, resulting in a benchmark comprising 11{,}967 data points.

\begin{figure}[H]
    \centering
    \includegraphics[width=\textwidth]{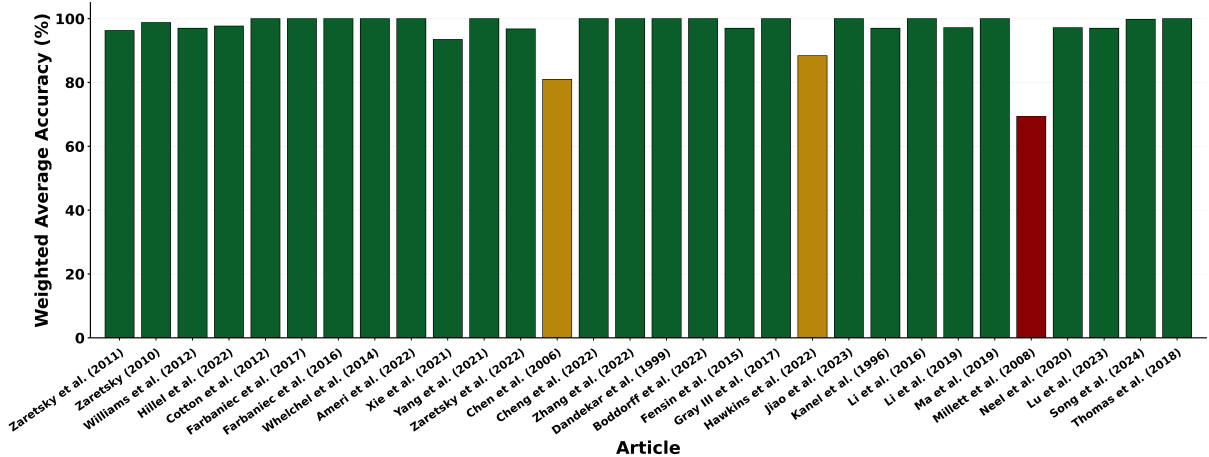}
    \caption{Weighted average extraction accuracy by article and aggregated at all extraction tiers using Gemini 3 Pro.}
    \label{fig:paper_accuracy}
\end{figure}

Figure~\ref{fig:paper_accuracy} shows the weighted average extraction accuracy by article for all 30 articles, aggregated across all extraction tiers (T1--T3). Overall performance is consistently high, with the vast majority of papers achieving accuracies above 90\%. Only three articles fall below this threshold~\cite{chenSpallBehaviorAluminum2006, hawkinsDynamicPropertiesFeCrMnNi2022, millettBehaviorNiNi60Co2008}. These lower-accuracy cases are mainly associated with heavy reliance on figure-based extraction (T3), where the LLM occasionally misinterpreted numerical values from images. Additional performance drops are observed when the workload exceeds approximately 500 extracted data points, because the model must backtrack across too many distributed values within a single article, which is a practical limitation of this approach. Despite the reduced performance on these three articles, the results still indicate strong and reliable extraction across diverse reporting formats and experimental conditions. The model successfully processed over a hundred individual shock experiments spanning multiple materials and loading conditions, producing structured output tables with hundreds of data entries and only minor errors. This consistency across numerous sources supports the robustness of the prompt-driven extraction methodology.

\begin{figure}[H]
    \centering
    \includegraphics[width=\textwidth]{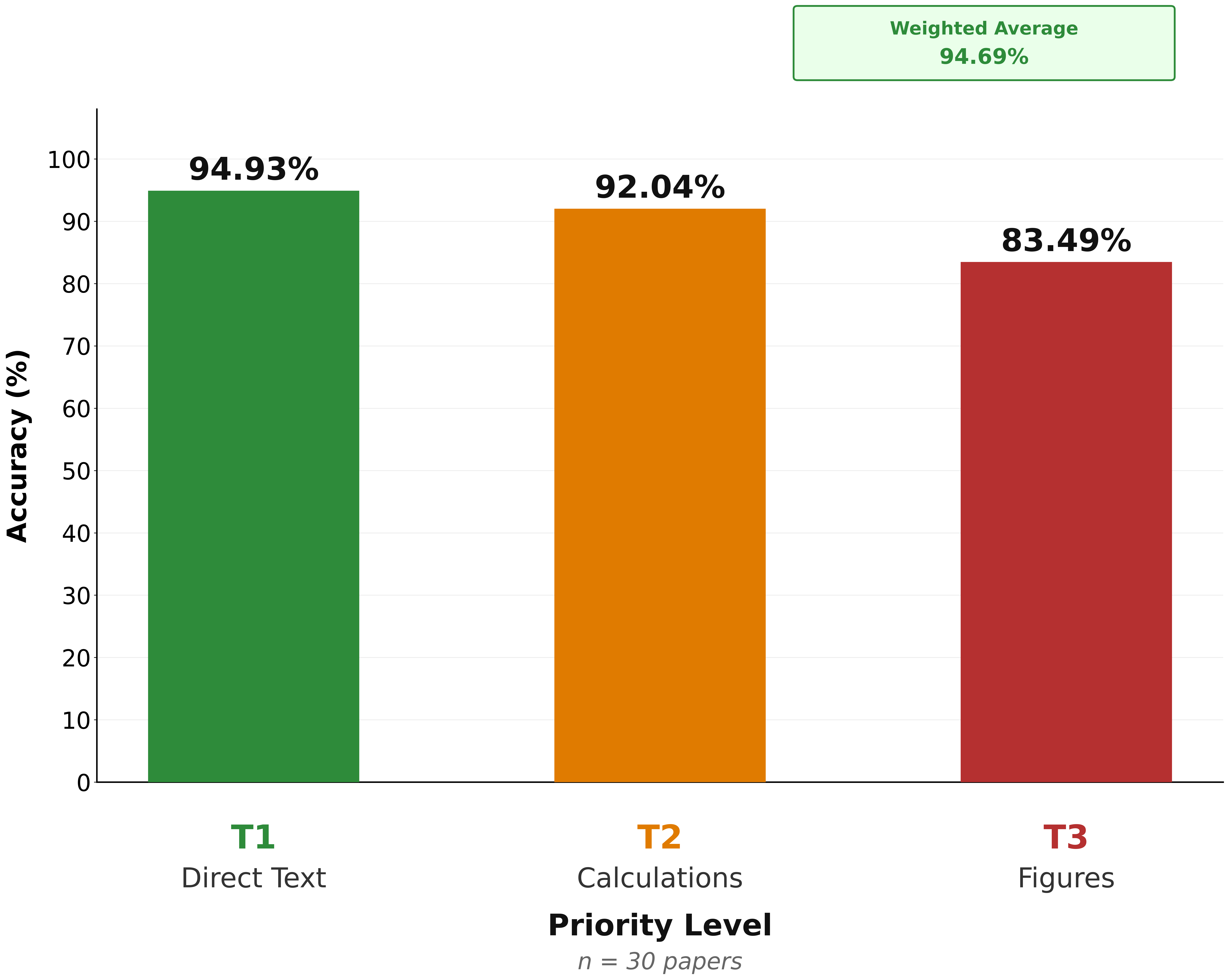}
    \caption{Accuracy for each extraction tier using Gemini 3 Pro for all articles combined. The overall weighted average is 94.69\%.}
    \label{fig:priority_accuracy}
\end{figure}

Figure~\ref{fig:priority_accuracy} shows the extraction accuracy by priority level, with individual bars reporting the performance for T1, T2, and T3, and a weighted average summarizing the overall performance across all tiers. The extraction achieved 94.93\% accuracy for T1, 92.04\% for T2, and 83.49\% for T3, with an overall weighted average accuracy of 94.69\%, weighted by the number of fields in each tier. The tier-wise distribution of extracted information also provides important context for interpreting these results. On average, T1 contains 328 data points per article, T2 contains 64 data points, and T3 contains 20 data points. This means that T1 dominates the total extracted volume and therefore contributes most strongly to the overall weighted accuracy, while T3 represents a much smaller but more challenging subset of the extraction task. The comparatively lower accuracy of T3 is therefore not only a reflection of its reliance on figure-based extraction, which is inherently more difficult than direct text or table extraction, but also of the fact that each error has a proportionally larger effect when the number of extracted data points is smaller. Taken together, these results indicate that the prompt-driven pipeline performs very strongly for directly reported and calculation-based fields, while still maintaining reasonable performance for the more difficult figure-derived entries.

It should be noted that the model processed every document in a single pass without requiring manual intervention during extraction, and all runs were completed without errors or timeouts, demonstrating both the accuracy and reliability of the automated pipeline.

It is important to note that the model demonstrated a marked absence of hallucinations, avoiding the fabrication of experiments or implausible physical values. These results suggest that the explicit constraints in Table~\ref{tab:prompt_rules} (specifically Rule D) effectively mitigated the generation of synthetic data. The observed errors were predominantly minor in nature and fell into two categories: (i) omissions, where a reported value was not extracted due to non-standard formatting or ambiguous textual presentation; or (ii) minor numerical inconsistencies arising from unit conversion or rounding. For example, in the Tier 2 extraction case reported by Farbaniec et al. (2017)~\cite{farbaniecSpallResponseFailure2017}, the flyer plate velocity was reported as $594$~m/s, whereas the LLM extracted a value of $590$~m/s. This corresponds to a relative error of $0.67\%$, which remains within the accepted tolerance for numerical matching. The discrepancy is likely attributable to intermediate rounding in the source document and does not materially affect the usefulness of the extracted dataset. 

Overall, the end-to-end extraction proved highly accurate and complete, demonstrating that LLM-based workflows can reliably handle structured data extraction tasks when guided by well-designed prompts. To further analyze the model's performance, in the following, we report the performance of each extraction tier.

\subsection{Tier 1 Extraction: Accuracy of Text and Table Extraction}



Tier 1 extraction (T1) is shown to be the most accurate and robust method, achieving 94.93\% accuracy. This is because the model handles explicitly reported information and therefore performs exceptionally well on clearly stated data. Across the 30 analyzed articles, Gemini 3 Pro accurately identified the key experimental parameters reported in the main text and tables. Out of a total of 9,835 extracted T1 data points, roughly 9,336 were extracted correctly.

The model also consistently handles partial or sparse data as instructed in the prompt. For example, in the article by Williams et al. (2012) \cite{williamsSpallResponse1100O2012}, where only three out of the ten discussed shots had a reported strain rate value, Gemini correctly records the three available values for their respective shots and inserts ``-'' for the remaining entries, whereas a naive approach would leave the entire column blank. This behavior demonstrates that the prompt effectively guides the model to be thorough while avoiding assumptions about data completeness across experiments. 

The model also preserves the formatting for uncertainties and enforces consistent units as specified by Rules~B (Uncertainty Extraction) and F (Unit Conversion) in Table~\ref{tab:prompt_rules}. For example, a yield stress recorded as $150 \pm 5~\mathrm{MPa}$ is output in that exact format, with the uncertainty notation preserved and units automatically converted to the canonical system (e.g., $150 \times 10^6 \pm 5 \times 10^6~\mathrm{Pa}$ in SI base units). 

The pipeline also successfully overcomes several extraction challenges through careful prompt design. For example, in the article by Neel et al. (2020) \cite{neelShockSpallLowalloy2020}, the shot identifiers are listed on one page, while the corresponding experimental conditions and measurements for those shots are reported on a different page. Rule~D (Experiment Identification) in Table~\ref{tab:prompt_rules} explicitly instructs the model to scan the entire document for all table continuations, to guarantee a successful resolution of this issue. This demonstrates the robustness of the prompt-driven approach in handling non-standard document layouts without requiring manual intervention.



Overall, the T1 track is shown to produce very high-quality extraction, effectively transcribing the source material's facts into the database with minimal loss or error.

\subsection{Tier 2 Extraction: Accuracy of Calculated Values from Equations}

Tier 2 extraction (T2) is an equation-driven extraction method, and it achieves 92.04\% accuracy, with roughly 1,765 correctly extracted values out of 1,918 total T2 data points. Although this is slightly lower than T1 extraction, it still demonstrates robust data extraction capability. When a required value is not explicitly stated in the text or tables, the model applies the governing equations in Table~\ref{tab:equations} to derive it from the other available quantities. For example, given $\sigma_{\text{HEL}}$ along with initial density $\rho_0$ and longitudinal sound speed $c_L$, the model computes the free-surface velocity $u_{\text{HEL}}$ through the relation $u_{\text{HEL}} = \sigma_{\text{HEL}}/(0.5\rho_0 c_L)$ and populates the corresponding field. It should be noted that the computed values of $u_{\text{HEL}}$ were consistent with other reported experimental magnitudes, on the order of $10^2~\mathrm{m/s}$ for metals, and passed the sanity-check ratio $\tau_{\text{HEL}}/\sigma_{\text{HEL}} \approx 0.25$--$0.3$.

Including these formulas in the prompt ensures that values not explicitly stated in the articles could still be computed from the other quantities reported in the same article. For example, elastic modulus $E$ and Poisson ratio $\nu$ are important material properties but are not always reported in articles on spall strength. When an article reports sufficient other properties (e.g., density and longitudinal and shear sound speeds), the model calculates and records $E$ and $\nu$ using the governing equations in the prompt. These computed values matched those obtained from the reported sound speeds and density and were consistent with properties reported from the literature. For example, for aluminum 1100-O, $E \approx 71.6~\mathrm{GPa}$ and $\nu \approx 0.33$ were obtained, aligning with reported values for this alloy \cite{williamsSpallResponse1100O2012}.

The drop in T2 accuracy relative to T1 stems from error compounding in multi-step derivations and occasional misidentification of input parameters. For example, in the article by Whelchel et al. (2014) \cite{whelchelSpallDynamicYield2014}, the model misread the upper HEL free-surface velocity for Shot~1323 as $u_{\mathrm{HEL},u} = 48~\mathrm{m/s}$ from the VISAR trace, while the value reported in the shot table is $u_{\mathrm{HEL},u}=54~\mathrm{m/s}$. Using the paper's $\rho_0=2.664~\mathrm{g/cm^3}$ and $C_L=6.367~\mathrm{mm/\mu s}$, the HEL relation $\sigma_{\mathrm{HEL}}=\tfrac{1}{2}\rho_0 C_L u_{\mathrm{HEL}}$ gives a value of $\sigma_{\mathrm{HEL},u}\approx 0.45~\mathrm{GPa}$, while the model's human-derived value was $\sigma_{\mathrm{HEL},u}\approx 0.41~\mathrm{GPa}$ (a $\sim 10\%$ deviation). However, iterative prompt refinement, particularly explicit unit-handling and numeric-transcription instructions (e.g., enforcing consistent SI units and cross-checking that computed moduli fall within physically reasonable ranges for the material), largely mitigated such errors.

Overall, T2 extraction successfully populates fields that would otherwise remain incomplete while maintaining minimal numerical discrepancies, confirming that appropriately instructed LLMs can reliably perform domain-specific calculations.

\subsection{Tier 3 Extraction: Accuracy of Figure Extraction}


Tier 3 extraction (T3), which is figure-based extraction, was the most challenging, yet achieving 83.49\% accuracy. This is the lowest reported across the three tiers but still surprisingly high. Reading values from plots is inherently approximate, and errors can arise from low-quality figures, difficulties in visual interpretation, or inaccuracies in assigning data extracted from a plot to the correct experimental set. Nevertheless, through the designed prompt, the workflow successfully extracts key data points from figures that would otherwise be omitted by other extraction approaches. For example, in the article by Chen et al (2006) \cite{chenSpallBehaviorAluminum2006}, which relied solely on a velocity--time profile to report the spall pull-back velocity $\Delta u_{\text{pb}}$, the model accurately measured $\Delta u_{\text{pb}}$ by identifying the decline in free-surface velocity following the peak, reporting a value of approximately $103 \text{ m/s}$ that matched manual digitization. 


Although the accuracy of T3 extraction is lower than that of T1 and T2, this result demonstrates that even complex visual data can be handled with reasonable reliability using prompt engineering. The accuracy of the extraction improved iteratively as the instructions were refined at each stage, which is consistent with prior work showing that expert-guided prompt refinement can improve scientific data extraction performance~\cite{liu_expert_grounded_prompt_2025}.

\begin{figure}[H]
    \centering
    \captionsetup[subfigure]{font=small, labelfont=normalfont, textfont=normalfont, justification=centering}
    \captionsetup{justification=justified, singlelinecheck=false}

    \begin{subfigure}[t]{0.44\textwidth}
        \centering
        \includegraphics[width=\textwidth, trim={0.1cm 0.8cm 0.7cm 0.8cm}, clip]{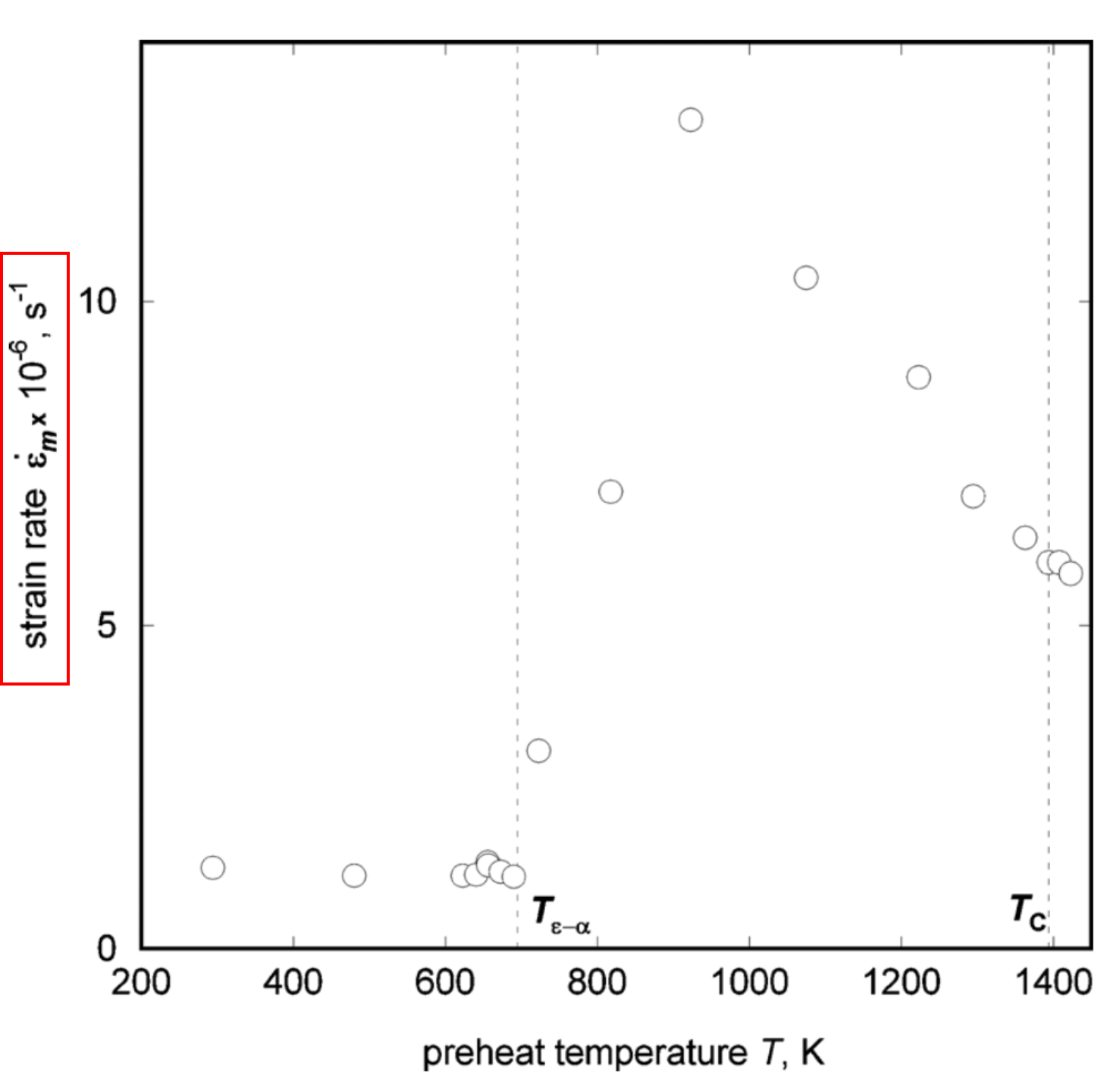}
        \caption{Axis multiplier misread}
        \label{fig:axis_multiplier_misread}
    \end{subfigure}\hspace{0.02\textwidth}%
    \begin{subfigure}[t]{0.48\textwidth}
        \centering
        \includegraphics[width=\textwidth, trim={0.6cm 0.6cm 0.6cm 0.6cm}, clip]{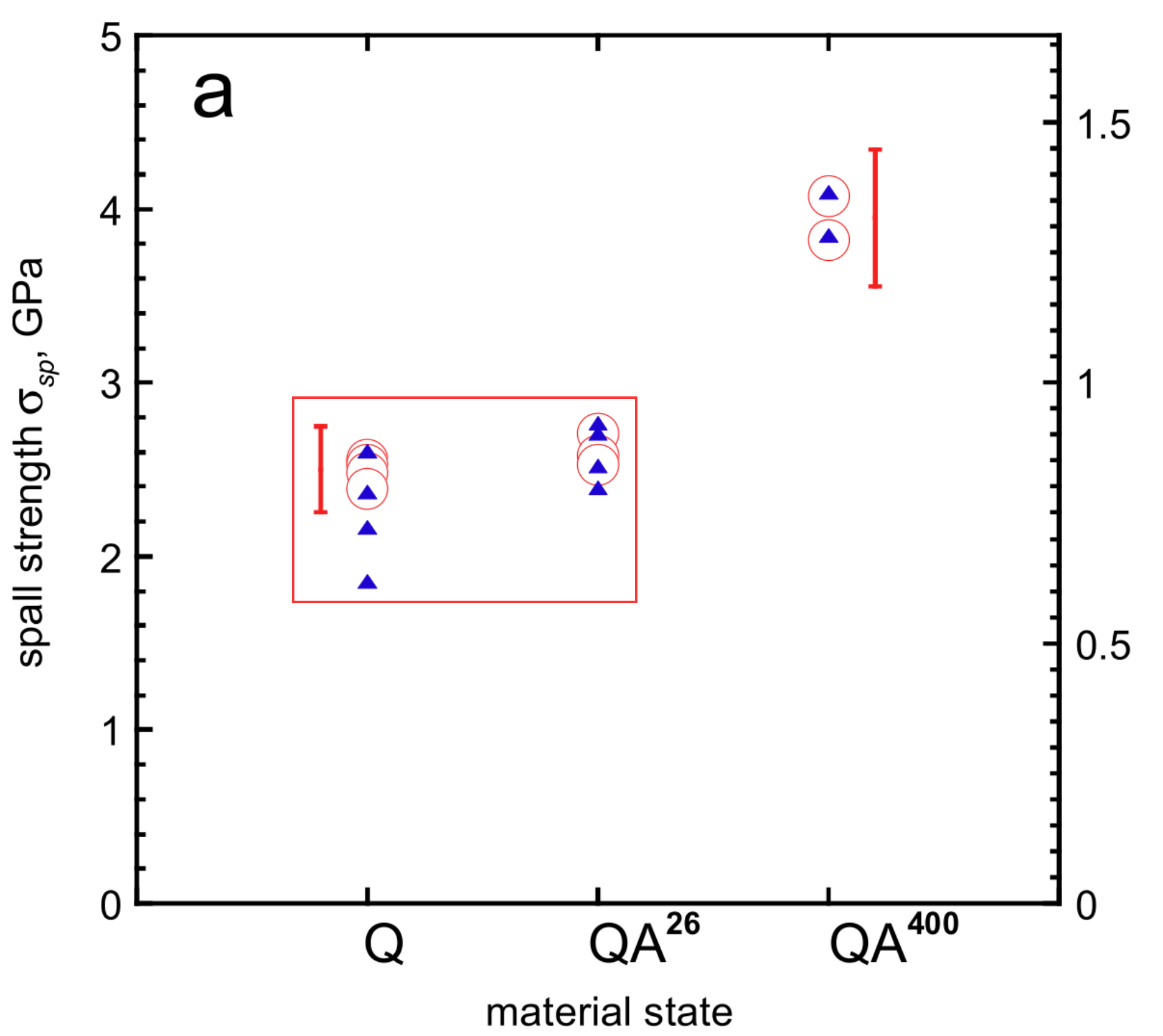}
        \caption{Marker overlap misassignment}
        \label{fig:overlap_marker_misread}
    \end{subfigure}

    \caption{Examples of figure-based extraction errors. (a) Small axis-title font can cause the LLM to misread the multiplier. Plot adopted from \cite{zaretskyImpactResponseCobalt2010} with permission. (b) Overlapping open-circle markers can cause the LLM to assign a plotted value to the wrong data point. Plot adopted from \cite{hillelShockWaveStudy2022} with permission.}
    \label{fig:t3_extraction_errors}
\end{figure}

The inaccuracies in T3 extraction are observed to arise from two sources: limited reading precision and ambiguity in mapping plotted data points to the correct experimental shot. For example, when extracting the strain-rate data from the appropriate figure in \cite{zaretskyImpactResponseCobalt2010}, which is shown in Fig.~\ref{fig:t3_extraction_errors}a, the model reported values multiplied by $1.0 \times 10^{5}\,\mathrm{s^{-1}}$ instead of the correct multiplier of $1.0 \times 10^{6}\,\mathrm{s^{-1}}$. This error was caused by the small font size, which led to the misreading of the axis multiplier. In a second example, when multiple data points are reported graphically without distinct labeling, as in \cite{hillelShockWaveStudy2022}, which is shown in Fig.~\ref{fig:t3_extraction_errors}b, the model assigned a spall-strength point to the wrong shot because overlapping open-circle markers made two experiments appear nearly identical. These errors highlight the inherent limitations of prompt-based figure interpretation and indicate that higher-quality images would likely improve model performance.



It should be emphasized that these kinds of errors occurred rarely in our analysis and could be systematically identified through post-analysis and sanity checks applied during prompt refinement. Although evaluation was performed with a strict numerical tolerance of 0.5\%, many of the T3 discrepancies remained small in practice and differed from the ground truth by only a few percentage points. Such deviations may remain acceptable for exploratory analysis or dataset construction, especially for figure-derived quantities, though further improvement could be achieved through enhanced figure-parsing protocols or hybrid human-in-the-loop validation. 

Overall, the T3 track provided valuable data that would otherwise have been missed, demonstrating the usefulness of LLMs for intelligent numerical extraction from figures.

\subsection{Extraction Performance Across Different LLM Platforms}


To assess the robustness of our approach across different LLM architectures, we used the same extraction prompt and articles with Claude Opus 4.5 (Anthropic's model) chat interface \cite{fazeli_claude_data_extraction_2025} and compared its performance to Gemini 3 Pro. 
We calculated the similarity of the data extracted from the two models using a closeness score ranging from 0 to 1, with 1.0 indicating that the models produced identical outputs for a given field (precisely the same value after normalization) and 0 indicating full divergence.
Divergence may arise from either (i) extraction errors by one or both models, or (ii) legitimate ambiguity in the source material where multiple valid interpretations exist. We define a closeness score as the fraction of fields for which the two models agree after unit normalization. An agreement for categorical fields requires an exact match, and numerical agreement uses the same tolerance rules as in the scoring protocol. The matching of missing entries (``--’’ in both outputs) is counted as agreement.

\begin{figure}[H]
    \centering
    \includegraphics[width=\textwidth]{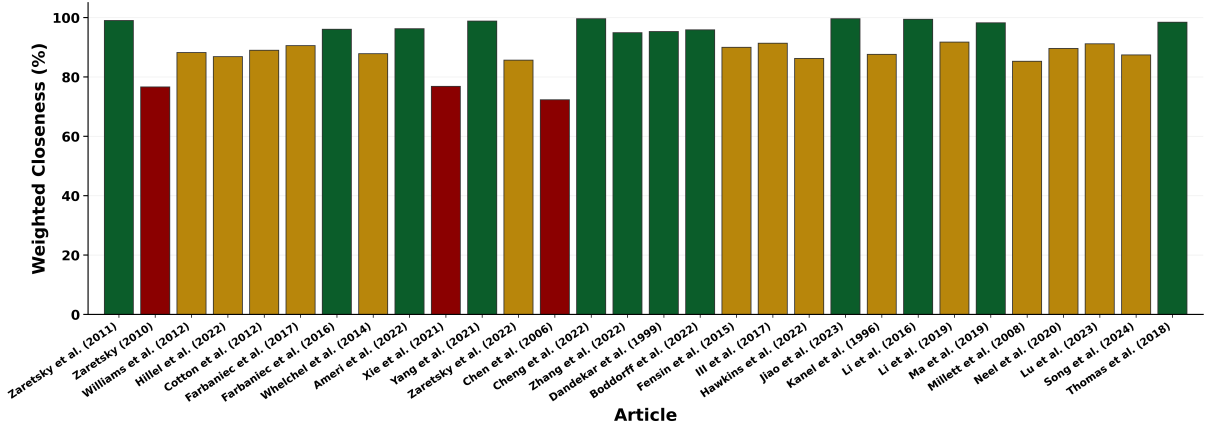}
    \caption{Weighted closeness score for Claude Opus 4.5 vs Gemini 3 Pro for all the papers.}
    \label{fig:closeness_paperwise}
\end{figure}

Figure~\ref{fig:closeness_paperwise} shows the closeness scores per article for all 30 articles, where each bar represents the weighted closeness between Claude Opus 4.5 and Gemini 3 Pro for each article. The majority of articles have a closeness score between 85\% and 97\%, indicating strong inter-model agreement and confirming that the prompt design successfully transfers across architectures. Nevertheless, three articles showed a lower closeness score, ranging from 70\% to 80\%. To understand whether these low closeness scores indicate genuine errors or legitimate interpretive differences, we manually reviewed cases with substantial divergence. The observed differences were primarily caused by three factors which we discuss in the following.


First, in some cases, one model extracted a value that the other completely missed. This type of divergence typically occurred when data were reported in non-standard formats or embedded within narrative text rather than structured tables. Such omissions reflect differences in the models' parsing strategies rather than fundamental disagreements about interpretation. For example, in the article by Hawkins et al. (2022)\cite{hawkinsDynamicPropertiesFeCrMnNi2022}, Gemini calculated the spall pullback velocity (m/s) using the T2 rule from known values, while Claude decided to omit it.


Second, some differences also arose from cases where the source material could reasonably be interpreted in more than one way. This revealed an important prompting requirement for ambiguous cases where more than one valid value is reported but only one must be entered. In such situations, the instructions should specify a default choice unless the paper clearly states that another value should be treated as the main one. For example, in the article by Whelchel et al. (2014)~\cite{whelchelSpallDynamicYield2014}, Gemini extracted the lower HEL value ($u_{\text{HEL},l}$) from a clearly labeled table column, while Claude selected the upper HEL value ($u_{\text{HEL},u}$) from the same table because it interpreted that value as the onset of yielding. Both choices can be justified from the way the table and text are written. This example shows that the difference came from ambiguity in the source material rather than from a clear model error, and it highlights the need to explicitly define a default selection rule in the prompt.

Third, inconsistencies also emerge in the way figure values were digitized or mapped from the plots. For example, in the article by Zaretsky et al. (2010)\cite{zaretskyImpactResponseCobalt2010}, Gemini and Claude extracted different spall strength values from the same figure, with Gemini reporting $0.45~\mathrm{GPa}$ and Claude reporting $0.55~\mathrm{GPa}$. Despite these discrepancies, the two models' outputs were quite similar in most domains, with an overall weighted closeness score of 88.76\% across all tiers of extraction.

\begin{figure}[H]
    \centering
    \includegraphics[width=\textwidth]{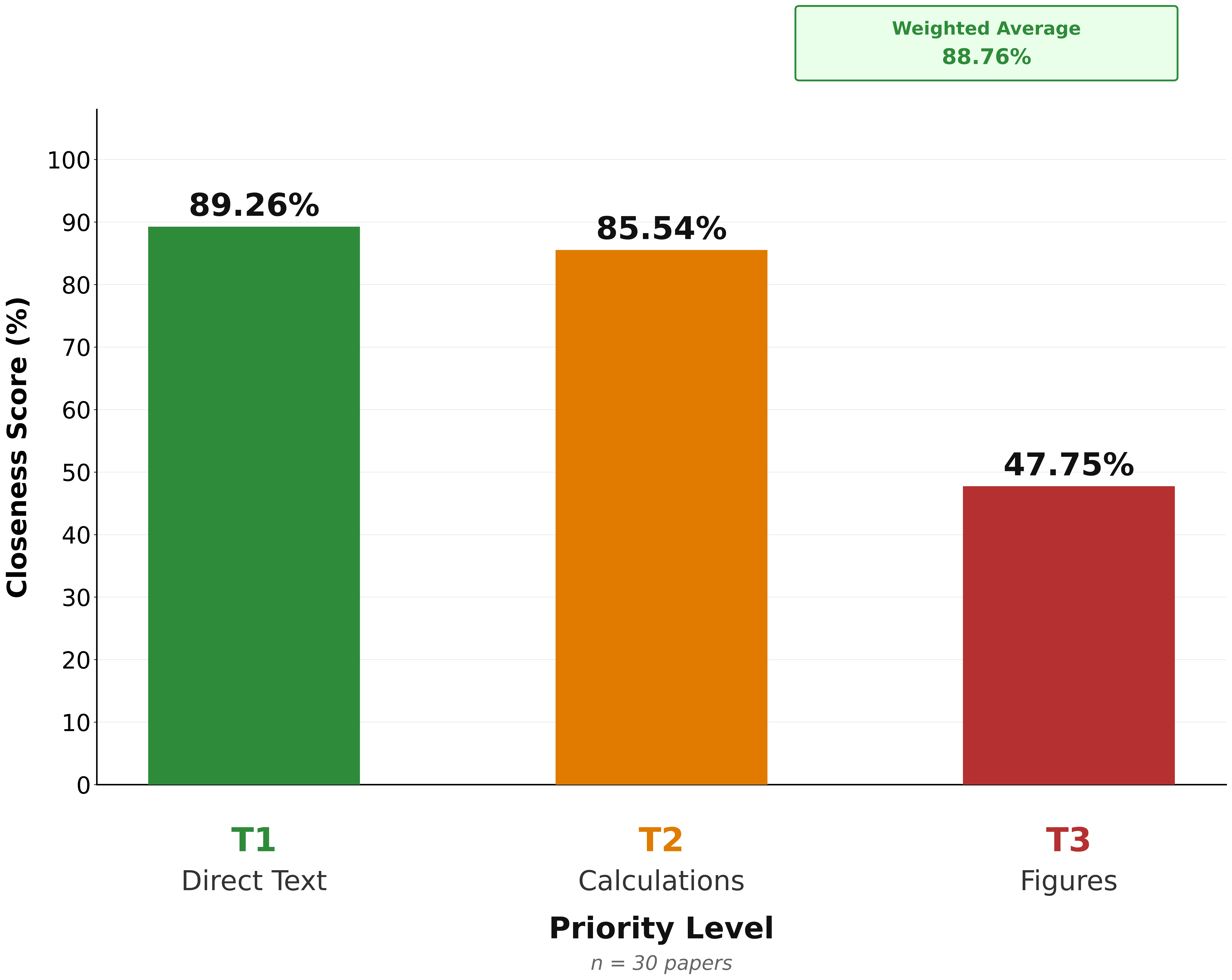}
    \caption{The closeness score for each extraction tier between Claude Opus 4.5 and Gemini 3 Pro. The weighted average across all three tiers is 88.76\%. This is based on data extracted from 30 articles.}
    \label{fig:closeness_priority}
\end{figure}

Figure~\ref{fig:closeness_priority} summarizes the closeness score between Claude~4.5 and Gemini~3\ Pro at each extraction tier. The bars represent the closeness score for T1, T2, and T3. Overall, T1 (Direct Text) extraction had the highest cross-model closeness, with a score of 89.26\%, indicating that both models almost invariably returned the same direct values for each field. This is a reassuring finding because it implies that when information is plainly present in the text or tables, advanced LLM models should extract it in a comparable manner. For T2 (calculated fields) extraction, the closeness score 85.54\%, indicating in a few cases one model computed a number that the other did not, or that there were minor numerical variations in the calculation logic between both models. The highest discrepancy was found in T3 (figure-based) extraction, where the closeness score dropped to 47.75\%.  

The results demonstrate that the prompt design supports high agreement between different models on structured extractions, especially for P1 and P2, whereas divergence in P3 reflects inherent challenges in interpreting figures by the two LLMs or ambiguous data reporting in the figures. The high-weighted closeness score across all 30 papers confirms that this protocol yields robust, reproducible outputs across model architectures and that careful prompt engineering can minimize model-specific variability during complex data extraction tasks.

\section{Discussion}\label{discussion}

\subsection{Prompting Principles}


Our results demonstrate that effective prompt engineering is critical to the success of LLM-driven extraction. The prompt used in this study is designed around four essential principles: clarity, context, structure, and iteration, all of which align with established best practices \cite {ateia_llm_information_extraction_2025}. 


Clarity ensures that the prompt specifies precisely what to extract, how to format the outputs, and how to handle missing or ambiguous values. In our implementation, we explicitly define the output structure (two tables with 37 columns (provided in Supplementary Materials Section S1.1\cite{supplement2026})), provide comprehensive symbol and abbreviation definitions, and enforce strict directives such as ``never invent data'' using unambiguous language. This explicit specification eliminates model uncertainty and ensures deterministic extraction behavior across repeated runs.


Context refers to embedding all relevant domain-specific knowledge and background information necessary for task execution within the prompt itself. We incorporate a comprehensive domain-specific context, including unit conversion tables, precise definitions of key terms (e.g., $\sigma_{\text{HEL}}$, $\tau_{\text{HEL}}$), and governing physical formulas (Table~\ref{tab:equations}), thus equipping the model with a self-contained knowledge base. This approach ensures that the extraction process relies on explicitly provided instructions rather than on the model's potentially outdated or incomplete parametric knowledge, thereby enhancing consistency and reducing reliance on implicit assumptions.


Structure indicates organizing the prompt in a logical and parsable manner that facilitates systematic processing. We partition our instructions into labeled sections with clear headings and enumerated rule sets, such as Rule A for unit conventions, Rule B for experiment identification, and Rule E for governing formulas, and include execution checklists for multi-step procedures. This hierarchical organization guides the model through the extraction workflow sequentially, much as a human annotator would be provided with a well-organized standard operating procedure that minimizes ambiguity and cognitive load.

Lastly, Iteration is the process of tuning the prompt (and the model's approach) via testing and feedback. We conducted several rounds of prompt tuning. In the first runs, flaws were highlighted (such as the model occasionally failing to convert units in calculations or not extracting partial data), which we fixed by adding specific rules and examples. We also used the model's own capabilities iteratively. For example, the prompt asks the model to perform a ``backtracking'' step (T2 computations after T1), a type of self-iteration within a single run.

\subsection{Implementing and Testing API Based Extraction}

In this study, 30 research articles were successfully converted into structured datasets with minimal human involvement using Gemini 3 Pro and a carefully engineered prompt. All 30 articles were PDF documents and were processed through the Gemini or Clauide chat UI, with each run producing the required output on the first attempt. To demonstrate scalability and transition the workflow toward high-throughput operation, the identical prompt was also tested through the Gemini API on five articles from the original set of 30 papers, namely \cite{cottonSpallStrengthNiobium2012, hillelShockWaveStudy2022, williamsSpallResponse1100O2012, zaretskyImpactResponseCobalt2010, zaretskyPlasticFlowShockloaded2011}. The API-based runs achieved 100\% accuracy, exceeding the performance observed with the chat UI. For the API runs, the Gemini Developer API was used with the selected model's default generation settings, with no explicit temperature, top-p, or top-k overrides, and one response was generated per document.

In the article by Hillel et al. (2022)~\cite{hillelShockWaveStudy2022}, the chat UI failed to capture the strain rate, whereas the API was able to identify the strain-rate equation from the article, extract the required parameters from the shot tables, and perform the necessary matching and calculations. The API can often perform better because it avoids hidden system instructions that chat interfaces may apply to control tone or style, which can sometimes interfere with strict extraction behavior~\cite{immanuel_object_oriented_llm_2025}. It also provides tighter control over generation settings, which helps the model remain factual and reduces hallucinations. In addition, the API offers greater control over the conversation context, reducing the likelihood that the core extraction instructions will be diluted or lost during the interaction.

A further extension of this workflow would be to use a multi-agent system in which different LLM APIs are used to fill gaps, cross-check uncertain fields, and resolve ambiguous entries. Such an approach could improve both completeness and accuracy, especially for difficult cases such as figure-based extraction or records with partial information. However, this added reliability comes with a clear trade-off in cost, since multiple model calls would be required for the same document and the total token usage could increase substantially. In practice, the choice depends on whether the priority is to minimize cost or to maximize accuracy, and this trade-off is an important consideration for scaling the pipeline to larger datasets.


\section{Conclusion}\label{conclusion}

This study demonstrates that carefully engineered prompts enable large language models to perform reliable, high-accuracy automated extraction of complex scientific data from literature. Using spall strength and related shock-physics parameters as a representative case study, we achieve over 92\% weighted accuracy across 30 research papers spanning diverse materials, experimental conditions, and reporting formats. Our approach successfully reconstructs complete experimental records by integrating three levels of extraction complexity: direct transcription from text and tables (P1, 94.93\% accuracy), equation-based derivation of missing fields (P2, 92.04\% accuracy), and figure-based digitization (P3, 83.49\% accuracy). Cross-model validation with both Gemini~3~Pro and Claude~Opus~4.5 yields 88.76\% inter-model agreement, confirming that the prompt design transfers robustly across different model architectures and produces consistent, reproducible outputs.

The primary contribution of this work is the demonstration of a structured, prompt-driven workflow for reconstructing complete experimental records from heterogeneous scientific literature sources. Through a three-level priority hierarchy embedded directly in the prompt, the workflow integrates information from text, tables, physics-based derivations, and figures, mirroring the decision logic of expert curators: first seeking explicitly reported values, then deriving missing quantities through physics-based calculations, and finally resorting to figure digitization only when necessary. This structure enables consistent, traceable reconstruction of experimental datasets, while preserving source tracability and enabling validation of derived quantities. By embedding domain knowledge and physical constraints directly into the prompt, the workflow enables general-purpose LLMs to perform domain-specific, physics-aware data reconstruction without requiring task-specific model fine-tuning or custom training pipelines, enabling single-pass processing of complete technical papers through a conversational chat interface accessible to researchers without computational expertise. This accessibility is critical since domain experts can adapt and refine both the hierarchical structure and the extraction protocol by iteratively modifying plain-language instructions rather than developing custom code or training specialized models.

The implications extend well beyond spall strength of alloys. Manual curation of scientific databases represents a persistent bottleneck across materials science, chemistry, biology, and engineering disciplines, where decades of experimental results remain locked in narrative text, heterogeneous tables, and figures. Automated extraction at 90+\% accuracy fundamentally changes the feasibility of large-scale meta-analyses, cross-study comparisons, and data-driven materials discovery. For instance, building comprehensive property databases that previously required years of manual effort can now be accomplished in weeks, enabling accelerated identification of structure--property relationships, validation of computational models against broader experimental evidence, and systematic exploration of underexplored parameter spaces in the published literature.

Moreover, this work demonstrates that structured prompt design can serve as a practical framework for encoding domain expertise into LLM-driven extraction workflows. The four-principle framework, including clarity, context, structure, iteration, provides a systematic approach for encoding expert knowledge into executable instructions. By progressively refining prompts through iterative testing and embedding physical constraints as verification rules, we convert implicit expert judgment into explicit, auditable extraction protocols that maintain traceability through priority tagging and evidence logging. This transparency and reproducibility are essential for scientific applications where data provenance and error characterization directly impact downstream analysis and decision-making.

In conclusion, this study demonstrates a practical, scalable pathway for leveraging LLMs to unlock decades of experimental knowledge embedded in scientific literature, transforming unstructured narrative into structured, traceable experimental datasets while maintaining the accessibility, transparency, and reproducibility required for rigorous scientific inquiry.


\section*{Acknowledgments}
Research was sponsored by the Army Research Laboratory and was accomplished under Cooperative Agreement Number W911NF-23-2-0062. The views and conclusions contained in this document are those of the authors and should not be interpreted as representing the official policies, either expressed or implied, of the Army Research Laboratory or the U.S. Government. The U.S. Government is authorized to reproduce and distribute reprints for Government purposes, notwithstanding any copyright notation herein.
\begingroup
\small
\bibliographystyle{abbrv}
\bibliography{references-2}
\endgroup
\end{document}